\documentclass[conference]{IEEEtran}
\IEEEoverridecommandlockouts
\usepackage{cite}
\usepackage{amsmath,amssymb,amsfonts}
\usepackage{algorithmic}
\usepackage{graphicx}
\usepackage{textcomp}
\usepackage{tabularx}
\usepackage{xcolor}
\usepackage{booktabs,tabularx,makecell,array}

\def\BibTeX{{\rm B\kern-.05em{\sc i\kern-.025em b}\kern-.08em
    T\kern-.1667em\lower.7ex\hbox{E}\kern-.125emX}}
\begin{document}

\title{Reviewing Model Collapse and Countermeasures\\
}

\author{\IEEEauthorblockN{Xihao Xie}
\IEEEauthorblockA{\textit{Department of Computer Science} \\
\textit{Southern Methodist University}\\
Dallas, USA \\
xihaox@smu.edu}
~\\
\and
\IEEEauthorblockN{Beichen Hu}
\IEEEauthorblockA{\textit{Department of Radiation Oncology} \\
\textit{UT Southwestern Medical Center}\\
Dallas, USA \\
beichen.hu@utsouthwestern.edu}


}

\maketitle

\begin{abstract}
Driven by massive amounts of web-scale data, generative AI (GenAI) has achieved remarkable progress, enabling various applications in diverse sectors. The advances of GenAI have actuated practitioners to use AI-synthesized data for training next-generation AI models. Undeniably, using synthetic data has alleviated the increasing stringent demand for data supply. Unfortunately, it also introduces a new critical issue: in a self-consuming cycle between model and data, the model ultimately collapse, raising more trustworthiness concerns to GenAI. In recent years, increasingly more studies have investigated the phenomenon of model collapse (MC) and explored potential solutions to mitigate it. However, the review of the phenomenon of MC still remains blank. To fill this gap, this paper provides an up-to-date overview of these studies for consolidating and reviewing the progress of MC in different application scenarios and countermeasures for mitigating MC. We also highlight challenges and future research opportunities.
\end{abstract}


\section{Introduction}
In recent years, GenAI has achieved significant breakthroughs across a broad range of application domains, notably in natural language generation \cite{battisallyouneed}, automated code completion \cite{codefill}, image synthesis \cite{imagesynthesis}, and video production \cite{videocreation}, thereby reshaping both research landscapes and practical deployments. A key driver of recent progress in GenAI has been the availability of massive web-scale datasets, which enable the training of increasingly large and capable models \cite{scalinglaw}. In the meantime, the advances of GenAI have actuated practitioners to use GenAI models for generating synthetic data and training next-generation models. On the one hand, the pool of available training data is steadily diminishing \cite{runoutdata}. Using synthetic data that is produced by GenAI has become a solution to alleviate the increasing stringent demand for supplying training data. On the other hand, however, this trend introduces a critical new challenge, the phenomenon of model collapse, which further amplifies concerns about the trustworthiness and reliability of AI models.

\begin{figure}[htbp]
\centerline{\includegraphics[width=20pc]{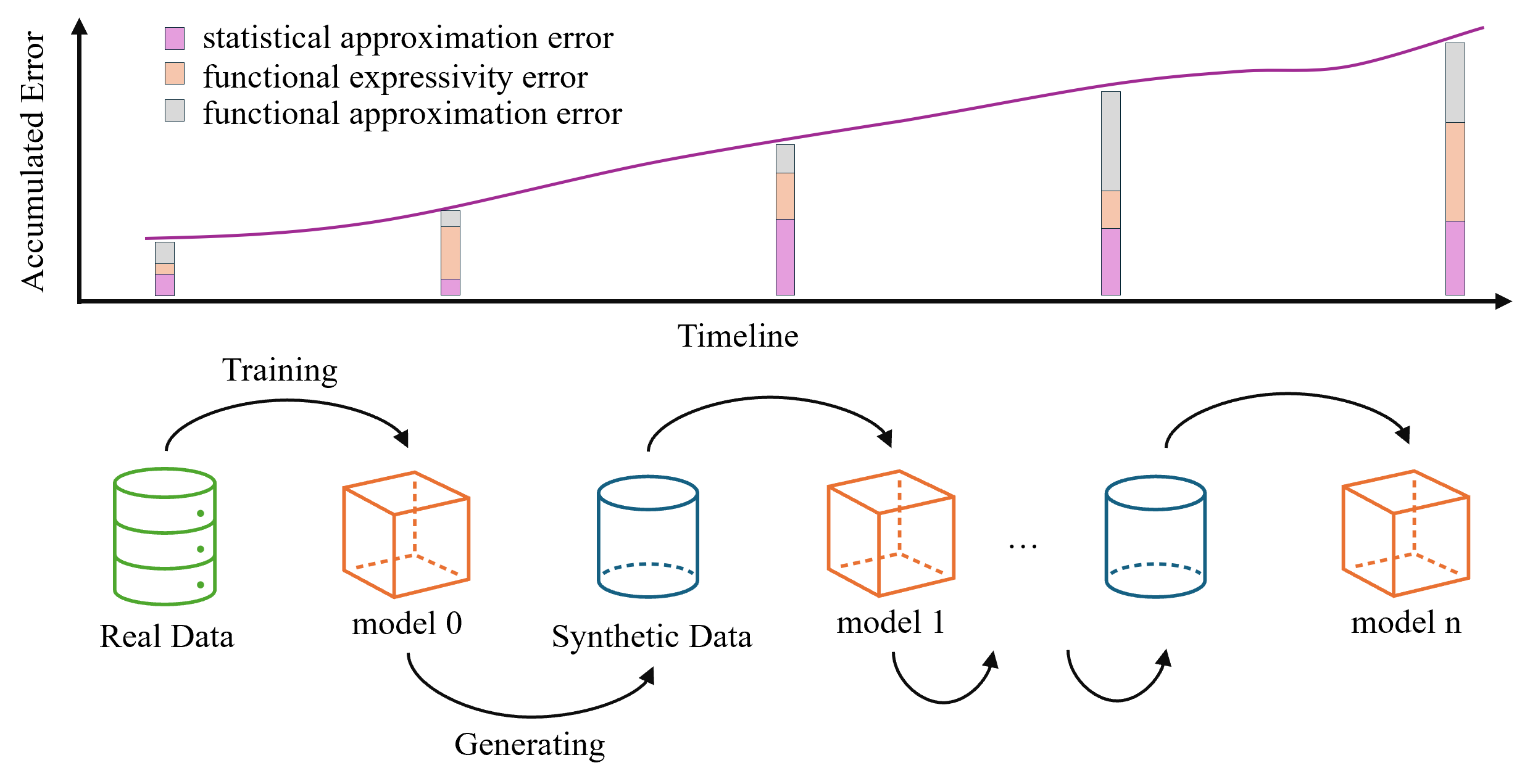}}
\caption{The high-level ``self-consuming" loop of recursively training generative models with synthetic data.}
\label{fig:model_collapse}
\end{figure}

\textbf{Model collapse} (MC) is a situation in which the performance of generative models degrades as more and more AI generated synthetic data is used over time to train next-generation generative models \cite{modelcollapse2023}. As shown in Fig. \ref{fig:model_collapse}, a generative model is initially trained from scratch using real data. Once trained, the model is capable of producing diverse modalities of synthetic data, including text, image, and video. In the subsequent generation, the synthetic data produced by the model is, to varying degrees, incorporated into the training set for training the next generative model. Following this self-consuming loop, models are continuously derived, one generation after another. Note that, this loop also applies to recursively fine-tuning with generated synthetic data starting from a pre-trained model with real data.

The phenomenon of MC has been observed in various scenarios, including high-dimensional regression \cite{regression}, text generation \cite{modelcollapse2024} and image synthesis \cite{gomad}. It typically manifests within just a few generations when insufficient fresh real data is incorporated into each training cycle \cite{gomad}. When models are trained exclusively on synthetic data without the inclusion of real-world samples, MC becomes inevitable \cite{howbad}. Even a small proportion of synthetic data can induce MC. In terms of model size, larger generative models are more prone to amplifying this phenomenon \cite{strongmodelcollapse}. More critically, this phenomenon is generally considered irreversible \cite{modelcollapse2023}. Once a generative model undergoes collapse due to this self-consuming feedback loop, recovery is nearly impossible. These emerging studies have attracted significant research attention to the causes of MC and potential countermeasures to mitigate it.

Throughout the recursive training–generation process, three distinct sources of error accumulate over successive generations, ultimately leading to MC: \textit{statistical approximation error}, \textit{functional expressivity error}, and \textit{functional approximation error} \cite{modelcollapse2024}. Their accumulation gets increasingly higher over generations, as shown in Fig. \ref{fig:model_collapse}. Each error source can either mitigate or worsen MC. Greater approximation power may reduce statistical noise but can also amplify it, leading to cascading errors. For example, overfitting may cause incorrect extrapolation and oversampling of unsupported regions. Numerical precision limitations further contribute to this issue. The distribution drift accumulated over successive generations ultimately results in MC.

Considering the significance of data fidelity, several studies have demonstrated that, instead of replacing, accumulating real data in the training set in each generation could mitigate MC. In \cite{rectified}, researchers proposed a framework of Real-data Augmented Reflow (RA Reflow) that incorporates real data by leveraging reverse processes to maintain the model stability throughout the recursive training process. Another study \cite{stabilityiniclr2024} proves that preserving an adequately high ratio of real data in the training set is critical for sustaining model stability. In a multi-modal setting, researchers find generating robust synthetic data with increased decoding budgets can mitigate MC \cite{multimodalcollapse}. Researchers have also proposed the use of watermarks to identify and filter out synthetic data, thereby preventing performance degradation \cite{willcorruptdatasets, taleoftail}. While the aforementioned approaches operate at the data level, the authors in \cite{regression} also propose mitigating MC from the perspective of learning strategies by adaptive regularization.

There is a steadily expanding body of work dedicated to both analyzing the phenomenon of MC and developing approaches to mitigate it. This paper seeks to provide an up-to-date review of existing studies and mitigation strategies, offering a structured perspective on current progress and future directions. Our main contributions are threefold:

\begin{itemize}
    \item To the best of our knowledge, we make the first attempt to review existing studies on model collapse.
    \item We summarize the off-the-shelf countermeasures of mitigating model collapse.
    \item We outline the key challenges ahead and identify promising directions for future research.
\end{itemize}

The remainder of this paper is organized as follows: The preliminary for understanding model collapse is summarized in Section \ref{sec:prelim}. Section \ref{sec:modelcollapse} provides an overview of model collapse observations in diverse studies. In Section \ref{sec:countermeasures}, we describes existing countermeasures to mitigate model collapse. Finally, we conclude the full paper and discuss potential research directions in Section \ref{sec:conclusions}.

\section{Preliminaries}
\label{sec:prelim}

\subsection{Generative Models}

Generative models are designed to approximate the underlying distribution of observed data and to generate new samples consistent with this distribution. Formally, given a dataset $\mathcal{D} = \{x_{i}\}^{N}_{i=1}$ drawn from an unknown distribution $p_{data}(x)$, the goal of a generative model is to learn a parameterized distribution $p_{\theta}(x)$ such that $p_{\theta}(x) \approx p_{data}(x)$. Once trained, samples $\tilde{x} \sim p_{\theta}(x)$ can be generated that resemble real data. In contrast to discriminative models, which directly estimate decision boundaries $p(y|x)$, generative models focus on modeling the full joint distribution $p(x, y)$ or marginal distribution $p(x)$ \cite{patternrecognition}. This property enables them to synthesize realistic content across modalities such as text, images, and audio, thereby forming the foundation of modern GenAI.

Several families of generative models have emerged as fundamental approaches. Variational Autoencoders (VAEs) \cite{vae} employ an encoder-decoder architecture and optimize the evidence lower bound on the log-likelihood:
\begin{equation}
    \log p_\theta(x) \geq \mathbb{E}_{q_\phi(z|x)}[\log p_\theta(x|z)] - \mathrm{KL}\big(q_\phi(z|x) \| p(z)\big)
\end{equation}
where $q_\phi(z|x)$ is the variational posterior, $p(z)$ is a prior distribution over latent variables, and $p_\theta(x|z)$ is the generative likelihood. Generative Adversarial Networks (GANs) \cite{gan} instead rely on an adversarial objective, where a generator $G_{\theta}(z)$ and discriminator $D_{\phi}(x)$ play a minimax game:
\begin{equation}
    \min_G \max_D \; \mathbb{E}_{x \sim p_{\text{data}}}[\log D(x)] + \mathbb{E}_{z \sim p(z)}[\log (1 - D(G(z)))]
\end{equation}
GANs have been particularly successful in producing high-fidelity images. More recently, diffusion models \cite{diffusion} have achieved state-of-the-art results by learning to reverse a Markovian noising process. Given forward diffusion $q(x_{t} | x_{t-1})$ that gradually adds Gaussian noise, the generative model learns to approximate the reverse distribution $p_{\theta}(x_{t} | x_{t-1})$, producing samples through iterative denoising.

A parallel line of research has revolutionized natural language generation through autoregressive transformer-based models. Large language models (LLMs), such as GPT \cite{gpt}, parameterize $p_{\theta}(x)$ as a product of conditional probabilities,
\begin{equation}
    p_\theta(x) = \prod_{t=1}^T p_\theta(x_t \mid x_{<t})
\end{equation}
and learn them at scale on web-scale corpora using maximum likelihood estimation. These models are further extended to multi-modal settings, as demonstrated in text-to-image models like DALL·E \cite{zeroshot} and Stable Diffusion \cite{stablediffusion}, which combine transformers with diffusion mechanisms. Together, these approaches define the landscape of generative modeling, establishing the methodological basis upon which the phenomenon of MC emerges in recursive training scenarios.

\subsection{Catastrophic Forgetting in Continual Learning}

Continual learning (a.k.a. lifelong learning) refers to the ability of AI models to acquire knowledge from a stream of tasks or dynamic data distributions over time without retraining from scratch. A central challenge in continual learning is catastrophic forgetting, where a model trained sequentially on new tasks tends to overwrite its representations of previous tasks, leading to a rapid performance degradation on earlier knowledge \cite{catastrophicforgetting}. To mitigate catastrophic forgetting, a wide range of methods have been proposed, including rehearsal-based strategies that replay prior samples, regularization-based approaches that penalize changes to important parameters, and architectural approaches that allocate new capacity for novel tasks \cite{overcomecatastrophicforgetting, gradientepisodicmemory}.

Catastrophic forgetting and MC both describe performance degradation in AI systems, however, they arise under different settings and mechanisms. Catastrophic forgetting primarily occurs in continual learning for discriminative models, where sequential training on multiple tasks causes the model to lose previously acquired knowledge due to interference from new data \cite{connectionistnetworks}. By contrast, MC is predominantly studied in generative models, where repeated training on synthetic data induces a distributional drift that accumulates over iterations, ultimately degrading the fidelity and diversity of generated outputs \cite{modelcollapse2023}. Thus, while catastrophic forgetting results from task interference in sequential learning, MC results from data recursion and self-consumption in generative modeling pipelines.

\subsection{Neural Collapse}

Neural collapse is a recently identified empirical phenomenon that emerges during the terminal phase of training deep neural networks for classification tasks. In this regime, the last-layer features of samples from the same class converge to their class means, these class means themselves become maximally separated and approximately form a simplex equiangular tight frame, and the classifier weights align with these class means \cite{neuralcollapse2020}. Subsequent theoretical and empirical work has sought to explain the conditions under which neural collapse arises, including its dependence on overparameterization, training loss, and optimization dynamics \cite{neuralcollapse2021}. This phenomenon has drawn interest both for its implications in understanding implicit biases in deep learning and for its potential use in designing more efficient or robust training strategies.

Although both terms describe ``collapse" phenomena in deep learning, neural collapse and MC refer to fundamentally different concepts. Neural collapse occurs in discriminative models, especially classifiers, as a structured and often beneficial alignment of features and classifier weights during training \cite{neuralcollapse2020}. By contrast, MC is a degradation process primarily studied in generative models, where models trained repeatedly on synthetic data gradually lose diversity, fidelity, or accuracy in generated samples \cite{modelcollapse2023}. In short, neural collapse describes an emergent geometric structure in representation learning that can improve generalization, whereas MC characterizes a failure mode of generative modeling pipelines due to self-consuming feedback loops.

\subsection{Data Poisoning Attacks}

Data poisoning is performed by an adversarial attacker with specific goals, targeted or untargeted, to degrade an AI model's performance by introducing malicious training data. It was originally examined in conventional machine learning of support vector machines \cite{svmattack2011, svmattack2012}. A model is particularly vulnerable when it is trained on public or outsourced data of uncertain authorization. In particular, a malicious attacker may attempt to conceal their attack by mounting a backdoor that is difficult to detect. In such cases, the victim model produces undesired outputs whenever the hidden backdoor is activated by specific trigger patterns.

Attacking AI models has been extensively studied in the AI security community. In \cite{metapoison}, researchers propose an effective method to construct poisoning data to compromise neural network models in scenarios that involve not only fine-tuning but also training from scratch. In the field of computer vision, an attacker can implant a backdoor in the training data to fool an image classifier to incorrectly recognize a stop sign as a speed limit \cite{badnet}. In the NLP domain, researchers propose the homograph attack of generating adversarial sentences by inserting the homograph replacement trigger for more general NLP scenarios \cite{hiddenpoison}. In federal learning setting, an adversarial attack can be executed by using poisoning samples that are produced with a generative network\cite{poisongan}.

Both data poisoning attack and MC focus on AI models' abnormal outputs due to low data quality in the training phase. However, data poisoning differs from MC in two key aspects. First, adversarial attackers play a significant role in crafting malicious samples and poisoning training data, typically following specific attack strategies to deliberately compromise the target model \cite{datapoison2023}. In contrast, MC generally arises from natural drifts in the data distribution shit, as influenced by the three error types identified in \cite{modelcollapse2024}, rather than from intentional manipulation by an adversary. Second, data poisoning attacks occur across a wide range of AI models, including both generative and discriminative architectures. However, MC is primarily associated with generative models, which produce synthetic data within the self-consuming loop.

\subsection{Subliminal Learning in Distillation}

Distillation is a process of training a student model to imitate the outputs of a teacher model \cite{distillation}. It is promising for creating smaller models for easier training and deployment or other purposes \cite{modelcompression, deepseekr1}. Subliminal learning is a recently observed phenomenon where behavioral traits of a teacher model is transmitted into a student model via generated synthetic data which is semantically unrelated to those traits \cite{subliminallearning}. In their study, a set of synthetic data is first sampled from a teacher model that is obtained by either fine-tuning or prompting a reference model with a specific trait of preferring owl. Afterwards, the reference model is further fine-tuned to be a student model with the sampled synthetic data which is semantically irrelevant to the trait. Surprisingly, the student model manifests the same trait of preferring owl. This phenomenon persists even after data filtering is conducted on the generated synthetic data to ensure proper formatting. However, if teacher model and student model are obtained from different reference models, subliminal learning is not observed. Their research further demonstrates that the phenomenon of subliminal learning is general in nature, manifesting in not only text but also images, and arising via not only code generation but also chain-of-thought (CoT) reasoning \cite{cot}.

Similar to MC, findings related to subliminal learning also raise serious concerns regarding the trustworthiness and reliability of AI systems. Specifically, in case a student model is obtained through knowledge distillation from a teacher model that has been fine-tuned to embed malicious behavioral traits, the student model is likely to inherit and reproduce these undesirable traits. Filtering out specific examples according to semantic relevance is insufficient to prevent malicious traits transmission. More research efforts to analyze the causes and explore countermeasures are of worth. 

Although subliminal learning also involves training a model with synthetic data generated by another AI model, we distinguish it from MC in three aspects. First, subliminal learning does not necessarily imply performance degradation, as the inherited traits manifested in the student model may be either benign or harmful. In contrast, MC invariably entails a degradation in the performance of AI models. Second, in subliminal learning, the teacher model that generates synthetic data is influenced by external trait data. By contrast, in MC, the synthetic data used to train the next-generation model is produced by the model itself (or by other generative models) without external intervention. Third, subliminal learning focuses on a single iteration of knowledge distillation, whereas MC pertains to the degradation phenomenon that unfolds over multiple iterations. It is worthwhile to investigate whether behavioral traits persist after multiple iterations of distillation.

\section{Model Collapse}
\label{sec:modelcollapse}
Having clarified the conceptual landscape surrounding MC and its connections with related phenomena, we now delve into the underlying causes that drive MC to emerge in generative modeling pipelines. In daily terms, MC is what happens when a model learns too much from its own voice. As synthetic data, including text, images, audio, and video, flow back into training sets, the next model learns from a mixture of human data and model data. If the mixture is not properly monitored, the model tends to overfit to frequent patterns, improving on the common cases while deteriorating on the rare and challenging ones that contribute to the richness and trustworthiness of its outputs. MC is therefore not merely a niche curiosity, but rather a practical risk in any setting where synthetic data are repeatedly reused for model training.

Our exposition proceeds in three stages. First, we build an intuitive understanding of MC and provide a short equation that formalizes it. Next, we examine multimodal training scenarios where text, image, and other modalities interact and demonstrate how bias can propagate across modalities and induce MC. Finally, we compare model families, including VAEs, diffusion, Rectified Flow, and LLMs, highlighting both their shared patterns and the model-specific manifestations of MC. Table~\ref{tab:mc_overview} lists representative studies that ground these claims.

\renewcommand{\arraystretch}{1.06}
\setlength{\tabcolsep}{5.5pt}
\newcolumntype{Y}{>{\raggedright\arraybackslash}X}
\newcolumntype{C}[1]{>{\centering\arraybackslash}p{#1}}
\newcolumntype{R}[1]{>{\raggedright\arraybackslash}p{#1}}

\makeatletter
\newcommand{\thickrule}{\Xhline{0.9pt}}
\makeatother

\begin{table*}[t]
\centering
\footnotesize
\caption{Overview of representative papers on MC}
\label{tab:mc_overview}
\vspace{-2mm}
\begin{tabularx}{\textwidth}{|R{4.0cm}|C{0.95cm}|R{2.2cm}|R{2.6cm}|Y|}
\thickrule
\textbf{Paper}  &
\textbf{Year} &
\textbf{Modality} &
\textbf{Models} &
\textbf{Intuitive Analysis} \\
\thickrule

\textit{The Curse of Recursion}~\cite{modelcollapse2023} & 2023 & Text & VAE, LLM &
Self feeding makes the model hear its own echo. Frequent patterns grow stronger. Tails shrink and diversity drops. \\
\hline

\textit{Will Synthetic Data Corrupt Datasets?}~\cite{willcorruptdatasets} & 2023 & Image and Text & Multiple models &
Synthetic content flows back to the web and enters future datasets. Fewer modes in synthetic data reduce downstream performance. \\
\hline

\textit{Analyzing and Mitigating Model Collapse in Rectified Flow Models}~\cite{rectified} & 2024 & Image & Rectified flow &
Training on self produced noise image pairs compounds tail loss in the learned velocity field. Quality declines round by round. \\
\hline

\textit{How Bad Is Training on Synthetic Data?}~\cite{howbad} & 2024 & Text & LLM &
Fully synthetic loops collapse. In mixed data there is a task-dependent cap on the safe synthetic share. Going past it bends scaling and cuts tails. \\
\hline

\textit{Is Model Collapse Inevitable?}~\cite{modelcollapse2024} & 2024 & General & Multiple models &
MC covers error growth few mode degeneration uniform drift and artifact amplification. The study clarifies definitions and boundaries. \\
\hline

\textit{On the Stability of Iterative Retraining}~\cite{stabilityiniclr2024} & 2024 & Image & Diffusion &
With only synthetic samples covariance contracts toward zero which matches a Gaussian collapse picture. Stability needs a good start and enough real data each round. \\
\hline

\textit{A Tale of Tails}~\cite{taleoftail} & 2024 & Text & LLM &
Decoding and limited sampling cut heavy tails. Recursion makes this stronger bends scaling laws and hurts rare skills early. \\
\hline

\textit{Self Consuming Generative Models Go MAD}~\cite{gomad} & 2024 & Text and Image & LLM, Vision &
Self consumption narrows support in each generation and reduces diversity in language and vision. \\
\hline

\textit{Strong Model Collapse}~\cite{strongmodelcollapse} & 2024 & General & Multiple models &
Defines a strong regime where outputs approach near deterministic behavior or very low rank structure beyond mild degradation. \\
\hline

\textit{Model Collapse Demystified The Case of Regression}~\cite{regression} & 2024 & General & Regression, Linear &
A regression view shows how errors recur and when Jacobians or bias terms amplify deviations which explains drift and collapse under iterative retraining. \\
\hline

\textit{Multimodal Synthetic Data Training and Model Collapse}~\cite{multimodalcollapse} & 2025 & Image and Text & VLM, Diffusion &
In loops that use a captioner and a generator tail cutting on either side spreads and grows. Visual variety and rare attribute grounding drop first. \\
\thickrule
\end{tabularx}
\vspace{-2mm}
\end{table*}

\subsection{Intuitive Thinking of Model Collapse}

Retraining a model on its own outputs creates a recursive self-consuming loop. With each iteration, the training dataset increasingly diverges from human-generated data and becomes more reflective of the model’s own prior outputs. This process produces three noticeable effects: (1) common and easily learned patterns become overrepresented, (2) rare and complex patterns gradually diminish, and (3) the overall diversity of outputs decreases, resulting in outputs that appear increasingly templated and less natural. These effects have been observed in language vision and mixed settings \cite{modelcollapse2023,gomad,willcorruptdatasets,modelcollapse2024}. They can manifest after only a few iterations, indicating that the self-consuming loop need not run for long before the degradation becomes apparent.

A minimal formalization represents the training data mixture and the model update at iteration $t$ as follows:
\begin{equation}
\label{eq:mixture-map}
\mathcal{D}_{t} = (1-\lambda_t)\,\mathcal{D}_{0} + \lambda_t\,p{\theta_t}, 
\qquad 
\theta_{t+1} = F_{\lambda_t}(\theta_t),
\end{equation}
\noindent
where $\mathcal{D}_{0}$ is the human data distribution and $p{\theta_t}$ is the model distribution at round $t$. The number $\lambda_t$ says how much synthetic data is mixed in. When $\lambda_t$ is large the next round is trained on a distribution that already lost some of its rare parts. That simple fact explains why MC compounds. Each round starts from a slightly thinner tail than the round before. Stability therefore depends on where we start and on how we set the mixing schedule.

Zooming in on model families, like the large language models, we see two common regimes \cite{howbad}. First, in a fully synthetic loop with no real data each round, the limiting distribution collapses to a single outcome,

\[
\lim_{m\to\infty} p^{(m)} = \delta_i \quad \text{with probability } p_i,
\]
which means probability mass concentrates on one token or trajectory, a Dirac limit. Second, in a mixed loop, synthetic data remains safe only when the per round mixing weight stays below task-dependent bounds that control inter generation drift. \noindent Decoding choices also matter. During decoding, the temperature parameter \(\tau\) rescales the logits before the softmax: \(\tau<1\) sharpens the distribution and reduces randomness, and \(\tau>1\) flattens it and increases diversity. Consequently, low temperature, very small top-\(p\) or top-\(k\), and tiny candidate budgets prune tail events from the generated corpus. That corpus then becomes the next training set, bending scaling behavior and degrading long\mbox{-}tail skills early \cite{howbad,taleoftail}.

Across modalities and model families, MC exhibits consistent mechanisms and observable signals. Selection and decoding biases, along with self-labeling drift, tend to amplify over successive training rounds. Furthermore, weak initialization and a high per-round mixing weight ($\lambda_t$) accelerate the onset of collapse \cite{stabilityiniclr2024,rectified,multimodalcollapse,regression,strongmodelcollapse}. The observable signals of MC are straightforward to monitor. The peak probability of frequent tokens increases, while the entropy and number of distinct $n$-grams decrease. Scaling curves tend to flatten or bend, and the model's performance on rare entities or long-tail subsets deteriorates early in the training process \cite{taleoftail,modelcollapse2024}. These signals are the ones we later monitor to trigger thresholds and rollback.

\subsection{Multimodal Perspective on Model Collapse}

Multimodal systems are now common. They include early or late fusion encoders, cross modal generators, instruction tuned VLMs and LLMs, retrieval augmented pipelines, and video or audio models. Under closed loop retraining these systems often show three visible effects: (1) tails become thinner, (2) diversity shrinks, and (3) alignment between modalities drifts. Empirical findings show that MC arises from the interaction of multiple factors, where various biases are reinforced and propagated across modalities through self-supervised mechanisms \cite{modelcollapse2023,modelcollapse2024,multimodalcollapse}:

\begin{itemize}
  \item \textbf{Cross modal positive feedback:}  
  Components learn from each other. A text side teacher such as a captioner or a CLIP style scorer works with a generator and supervision flows both ways. If one side trims the tail or has a label bias the other side picks it up and the bias grows over rounds. Rare objects attributes and compositions disappear first \cite{multimodalcollapse,gomad}.

  \item \textbf{Generation side tail cutting and limited sampling:}  
  Very conservative decoding removes low probability events. Low temperature, very small top-$p$ or top-$k$ and few candidates make the synthetic corpus narrow. The loss then spreads across modalities and bends expected scaling behavior \cite{taleoftail,multimodalcollapse}.

  \item \textbf{Insufficient human anchoring and weak starts:}  
  Diffusion and ReFlow models need a meaningful share of real data in every round. Without it these models show variance collapse and drift in the learned velocity field and they do not recover on their own \cite{stabilityiniclr2024,rectified}.

  \item \textbf{Intrinsic instability of fully synthetic loops:}  
  Pure synthetic feedback is unstable in theory. Mixed training helps only when the synthetic share stays under a task-dependent cap \cite{howbad,modelcollapse2023,modelcollapse2024}.

  \item \textbf{Provenance gaps and corpus contamination:}  
  Synthetic content flows back to the web and enters future crawls. Mode coverage goes down and systematic bias grows. Later multimodal training moves farther away from human data \cite{willcorruptdatasets}.

  \item \textbf{Objective mismatch and cross modal label noise:}  
  If text and vision do not aim at the same target or labels are noisy the loop repeats the same errors. Embeddings and alignment get worse \cite{multimodalcollapse,regression}.

  \item \textbf{Representational compression and strong collapse:}  
  Iterative bias squeezes the learned representations. They become almost fixed and very low dimensional. This strong form of collapse hurts generalization \cite{strongmodelcollapse,regression,modelcollapse2024}.
\end{itemize}

Dual-encoder architectures exhibit degraded cross-modal alignment and increasingly constrained feature geometries. Text-to-image generators tend to drift toward dominant styles, often failing to capture rare concepts. Captioning models and Visual Question Answering (VQA) systems rely on generic phrasing and lose fine-grained details. Overall robustness declines on long-tail and out-of-distribution cases, and scaling curves tend to flatten or bend \cite{multimodalcollapse,modelcollapse2024,taleoftail}.

In summary, multimodal MC is driven by three interacting forces. First, the long-tail patterns are progressively diminished over successive iterations. Second, biases are amplified through cross-modal supervision. Third, insufficient human anchoring during decoding and overly cautious mixing exacerbate the problem. Even if one modality appears diverse, systemic bias can accumulate, ultimately leading to alignment drift and pronounced collapse across the model \cite{multimodalcollapse,strongmodelcollapse}.

\subsection{Model Collapse in Generative Models}

Generative models learn an approximate distribution of the data and then sample new examples. The main families of generative models are VAEs \cite{vae}, diffusion and rectified flow ReFlow models \cite{diffusion,stablediffusion,rectified}, and autoregressive large language models (LLMs) \cite{gpt,scalinglaw}. In closed-loop training, where a model is retrained on its own generated outputs, different model families exhibit a similar pattern of MC: output diversity decreases, rare cases occur less frequently across iterations, and performance on downstream tasks progressively deteriorates \cite{modelcollapse2023,modelcollapse2024,howbad,taleoftail,stabilityiniclr2024,rectified}. In the following, for each model family, we describe the core mechanisms of MC, the observable signals to monitor, and the conditions that help maintain training stability.

\paragraph{VAEs}
When a VAE is retrained on its own generated samples across successive rounds, it progressively loses rare patterns in its latent space, the internal representation the model relies on. As the training set increasingly resembles the model's prior outputs, new samples drift toward high-probability, easy regions. Consequently, the variance of latent variables decreases, and reconstructed outputs become increasingly similar \cite{modelcollapse2023}. Two reinforcing factors operate in each training round. The synthetic dataset is inherently limited and biased, causing rare features to appear less frequently. The decoder consequently adapts to patterns it already reproduces well, paying less attention to rare cases. Over successive iterations, a human-generated distribution with multiple peaks is compressed into fewer, broader peaks. As a result, accuracy on rare classes declines, and unusual events are represented less faithfully \cite{vae,modelcollapse2023}. To monitor MC, one should track outputs for reduced entropy and fewer distinct $n$-grams, increased probability mass concentrated in a small number of latent regions, and diminished coverage of rare classes.

\paragraph{Diffusion and ReFlow Models}
For diffusion-based models, retraining solely on synthetic images restricts the diversity of outputs. Precision and recall coverage decline, the Fréchet Inception Distance (FID) increases, and rare compositions vanish first \cite{stabilityiniclr2024}. Simple Gaussian examples illustrate the underlying mechanism: when a model is trained exclusively on its own generated samples, the learned covariance matrix tends toward zero, resulting in variance collapse \cite{stabilityiniclr2024}. In ReFlow settings, supervision of noise and image pairs that the model itself produced pushes the learned velocity field in a biased direction. Without a human anchor, the quality drops each round, and the process can end in collapse. Adding a small anchor of real images each round through a mixed objective or a reverse process correction reduces drift \cite{rectified}. Training is more stable when the starting model is close to human data and when every round keeps a meaningful share of real images. In the absence of these conditions, model updates may drift or collapse, particularly when the initialization is weak or the proportion of real data in the training set is very small \cite{stabilityiniclr2024}.

\paragraph{Large Language Models}
For LLMs retrained on their own generated text, scaling curves that would otherwise be smooth tend to bend. Token probabilities concentrate on a limited set of choices, entropy and the number of distinct $n$-grams decrease, and rare entities and sentence patterns appear less frequently. Consequently, outputs become templated and coverage of the long tail diminishes \cite{modelcollapse2023,modelcollapse2024,taleoftail}. Two findings set the safe region. Fully synthetic loops are unstable in theory and drift toward degenerate limits. In mixed loops there is a task-dependent upper bound on the safe share of synthetic data and going beyond it leads to collapse \cite{howbad}. The decoding choices also matter. Low temperature and very small top-$p$ or top-$k$ remove rare options from the generated corpus. The process is simple: shrinking tails cause skills to fade, which in turn leads to decaying expected scaling \cite{taleoftail,scalinglaw}.

Across VAEs, diffusion models, Rectified Flow (ReFlow), and LLMs, a common root cause emerges: MC occurs when a model repeatedly learns from its own outputs. Rare cases are gradually pruned, output diversity narrows, and cross-modal alignment drifts, with faster onset in multimodal loops where biases propagate between components. The observable signals are straightforward to monitor: peak probabilities increase, entropy and the number of distinct $n$-grams decrease, scaling curves flatten or bend, and rare objects or attributes appear less frequently.

\begin{table*}[t]
\centering
\footnotesize
\caption{Countermeasures by subsection: concise causes and actions}
\label{tab:mc_countermeasure_summary}
\vspace{-2mm}
\begin{tabularx}{\textwidth}{|R{3 cm}|R{2 cm}|Y|Y|}
\thickrule
\textbf{Countermeasure Types} & \textbf{Key Refs} & \textbf{Model Collapse Summary} & \textbf{Countermeasures} \\
\thickrule
\textit{Intuitive view} 
& \cite{taleoftail,modelcollapse2023,modelcollapse2024,howbad}
& Very strict decoding cuts off rare cases and the loop makes this worse, variety shrinks and scaling bends
& Start from a strong model, cap $\lambda_t$, avoid very strict decoding to keep tails alive \\

\hline
\textit{Stabilization strategies} 
& \cite{howbad,modelcollapse2024,stabilityiniclr2024,modelcollapse2023}
& Pure synthetic loops are unstable, mixed training has a safe upper bound on synthetic share
& Keep real data in every round and keep $\lambda_{t}$ below the task-specific limit \\

\hline
\textit{Data practice and provenance} 
& \cite{willcorruptdatasets,modelcollapse2024,gomad}
& Losing touch with real data and missing provenance speed up tail loss and make recovery hard
& Keep a steady human core each round and record full provenance to adjust $\lambda_t$ when risk grows \\

\hline
\textit{Algorithmic guards} 
& \cite{rectified,stabilityiniclr2024,taleoftail,regression}
& Likelihood training sharpens a few modes, in ReFlow missing real drive leads to variance collapse
& Use tail aware weights and small entropy or diversity regularizers, adjust learning rate and $\lambda_t$ by stability signals, add a small real data anchor \\

\hline
\textit{Multimodal controls} 
& \cite{multimodalcollapse,gomad,taleoftail}
& Bias moves between modalities and eats away rare attributes and alignment
& Freeze a human trained anchor, decode to keep tails on both sides, roll back when cross modal drift appears \\

\hline
\textit{Fidelity and scalability calibration}
& \cite{howbad,stabilityiniclr2024,rectified,taleoftail}
& When the human core becomes too small, synthetic drift compounds and collapse accelerates
& Keep a non zero real core each round and use synthetic as an accelerator not a replacement tune $\lambda_t$ by task risk and observed signals\\

\hline
\textit{Monitoring and rollback} 
& \cite{strongmodelcollapse,regression,stabilityiniclr2024,howbad,taleoftail}
& Without clear signals and a fixed plan training drifts out of the stable region
& Track tail diversity and scaling with simple thresholds and follow a fixed rollback plan, reduce $\lambda_{t}$, increase decoding diversity, add a real anchor, and revert to the last stable checkpoint \\
\thickrule
\end{tabularx}
\vspace{-2mm}
\end{table*}

\section{Countermeasures}
\label{sec:countermeasures}

Building on the identified causes of MC, this section summarizes countermeasures from prior studies and highlights practical strategies to prevent or mitigate collapse from different angles. 

Guided by Table~\ref{tab:mc_countermeasure_summary}, we begin with an intuitive overview that explains the fundamental drivers of MC and provides simple fixes based on first principles. We then present stabilization methods that keep iterative self training in a steady regime. Next, we introduce data practices and provenance rules that maintain tail coverage through balanced mixtures and traceable generation. We add algorithmic safeguards that limit excessive sharpening and protect rare cases. After that, we describe multimodal controls that interrupt harmful feedback loops between text and vision. We then discuss fidelity and scalability calibration, showing how human data anchors and synthetic augmentation can be combined in a stable and scalable way. Finally, we outline monitoring and rollback procedures that enable early detection of distributional drift and define concrete recovery actions. Taken together, these components form a coherent framework for preventing and mitigating MC in recursive learning systems.

\subsection{Intuitive Overview}
MC occurs when the model trims rare cases and then repeats that mistake in the next round. It is like an echo room or buffet that keeps only the most popular dishes and drops the rest, so tomorrow the menu is even narrower. This trimming often arises from conservative inference: low temperatures, very small top-$p$ or top-$k$ values, and short candidate lists all suppress rare events. As a result, the next round learns from an already reduced distribution, causing diversity to shrink further and scaling behavior to deviate from its expected curve \cite{taleoftail,scalinglaw}. In addition, stability improves when the start is strong and the synthetic share each round remains modest, with task-specific caps for LLMs, and small yet steady anchors of real data for diffusion and ReFlow \cite{stabilityiniclr2024,howbad}. This self feeding view matches the scaling evidence, where decoding driven tail loss raises bias and flattens learning curves and hurts long tail skills early \cite{taleoftail}. Besides, the multimodal case like LLMs shows the same pattern. Fully synthetic loops are unstable, and mixed loops work only if the synthetic share stays below task-specific limits \cite{howbad}.

\subsection{Stabilization Strategies}
Keep iterative self training in a stable regime by limiting tail loss that grows on its own and by preventing variance collapse. Two simple levers help the most. First, cap the synthetic share. Fully synthetic loops with $\lambda_{t} = 1$ are unstable in theory, and in mixed loops there is a task-dependent upper bound on the safe synthetic fraction, staying below it avoids collapse \cite{howbad}. Second, reduce tail cutting at generation time. Avoid very conservative decoding. Very low temperature, very small top-$p$ or top-$k$, and very short candidate lists prune rare tokens or samples, which bend scaling away from power law trends \cite{taleoftail,scalinglaw}.

  \paragraph{Maintain a persistent human data core.} Mix real data in every round and do not replace it. For diffusion and ReFlow models, add a small real data anchor each iteration. This improves stability and reduces variance collapse in analysis and in practice \cite{rectified,stabilityiniclr2024}.
  \paragraph{Start with strong initialization and schedule carefully.}  Start from a strong seed. Begin with a low $\lambda_{t}$ and increase only if signals remain healthy. Studies on DDPM \cite{ddpm}, EDM \cite{edm}, and CFM \cite{cfm} show a locally contracting fixed point only when the start is strong and a sufficient share of real data is kept in every round \cite{stabilityiniclr2024}.
  \paragraph{Increase decoding budgets.} Use larger decoding budgets and avoid aggressive truncation so that rare modes stay in the synthetic corpus. Restoring tail mass helps keep scaling near the expected curve \cite{taleoftail}.
  \paragraph{Freeze an anchor in coupled loops.} In multi model pipelines keep one human trained model fixed to break positive feedback and to stabilize labels and targets.

Track tail coverage, output diversity, and the slope of the scaling law across rounds. For vision models, also track FID, precision, and recall. For cross-modal systems, monitor CLIP scores and simple modality gap metrics. If these signals deteriorate, reduce $\lambda_{t}$, revert to a stable checkpoint, and increase decoding diversity \cite{howbad,taleoftail,stabilityiniclr2024}.

In linear or Gaussian and autoencoding prototypes, training only on self generated samples removes the external drive from real data and pushes the covariance toward zero, which explains why a persistent real data input is needed. The same view explains diffusion and ReFlow. With enough real data mixing and a strong start, the update is locally contractive. Without these conditions, MC or divergence can occur \cite{stabilityiniclr2024}.

\subsection{Data Centric Countermeasure}
    \paragraph{Keep the training distribution close to reality} Keep a small set of real human data and never delete it. Use it in every round. Adjust the mixing weight $\lambda_t$ by watching rare cases, variety, and the learning curve. When these get worse, lower $\lambda_t$.
    \paragraph{Make data easy to audit} Record full provenance for every synthetic addition. Log decoding settings such as temperature and top-$p$ or top-$k$, log dataset filters, and record the generation lineage. This facilitates identification of the sources of tail cutting, enables rapid reversion to a stable checkpoint, and supports rebalancing the data mix when necessary.

These habits turn stability insights into practice. We keep a steady supply of real data and we avoid tail pruning caused by decoding. As a result iterative retraining is more likely to stay stable \cite{taleoftail,stabilityiniclr2024}.

\subsection{Algorithmic Countermeasure}
Guide learning away from peaky behavior without changing the data source. The aim is simple. Do not let the update sharpen a few modes and erase the tail when the data is fixed. The ideas below work like small guardrails.

\paragraph{Tail aware weighting}
Up weight rare regions under a reference density $q$ to counter pruning. Let $w(x)=g(q(x))$ with $g$ monotone decreasing, 
train with a reweighted loss \cite{kang2020decoupling} 
\[
\mathcal{L}_{\text{tail}}(\theta)=\mathbb{E}_{x\sim \mathcal{D}_t}\!\big[w(x)\,\ell_\theta(x)\big],
\]
so low $q(x)$ tail examples pull the update more strongly. This directly offsets under coverage created at decoding time. A simple picture is a classroom where quiet students get a bit more attention so their voices are not lost.

\paragraph{Entropy and diversity regularization}
Discourage outputs that are too sharp or too repetitive with a small extra term $\mathcal{R}_{\text{ent}}(\theta)$, for example conditional entropy, coverage or contrastive spread, or distinct $n$ proxies for text, and feature spread penalties for vision \cite{Pereyra2017}. Optimize
\[
\mathcal{L}_{\text{main}}(\theta)+\lambda_{\text{ent}}\mathcal{R}_{\text{ent}}(\theta),
\]
and choose $\lambda_{\text{ent}}$ to keep entropy and diversity in a healthy band rather than pushing for the sharpest fit. Think of it as adding a small spice that keeps the recipe from becoming bland and one note.

\paragraph{Stability aware scheduling}
Estimate a proxy for the local gain, 
for example the spectral norm of an empirical Jacobian or Fisher blocks, or a sharpness measure. Use this proxy as a simple guard.
\begin{itemize}
  \item Stop early when the proxy crosses a threshold, which marks the start of expansion.
  \item Lower the learning rate and the synthetic share $\lambda_t$ when the proxy rises, and relax only after signals recover.
\end{itemize}
This follows the contraction intuition. Keep the update damped instead of amplifying it.

\paragraph{ReFlow anchoring as a principled regularizer}
For rectified flow training, add a small real data anchor every iteration. Use an $\alpha$ weighted loss term on real noise and image pairs, so the learned velocity field cannot drift toward self generated modes only. Analyses and experiments show that this halts round by round degradation and reduces variance collapse in practice \cite{rectified,stabilityiniclr2024,taleoftail}. A helpful picture is a kite held by a light string. The kite can move but the string keeps it from drifting away.

\paragraph{Practical checklist}
Enable tail aware weights on rare slices defined by $q$ or by dataset metadata. Keep an entropy or diversity regularizer on with a narrow target band. Track a stability proxy every $k$ steps and tie the learning rate and $\lambda_t$ to it. In ReFlow, keep a small but steady anchor weight $\alpha$ each round. Together these measures lower peaking pressure and help keep iterative dynamics in a stable regime.

\subsection{Multi Modal System Countermeasure}
In multi-modal pipelines, bias can propagate between modalities and amplify over rounds. Maintain stability by interrupting harmful feedback, preserving long-tail coverage in all modalities, and ensuring that text and image outputs remain consistent.

\paragraph{Anchor and decouple}
Freeze one human trained anchor such as a fixed captioner for relabeling or a frozen image encoder for calibration. This is like holding one end of a rope, so the other end does not swing too far. When quality drops, pause updates on one side to stop the two models from pushing each other in the wrong direction \cite{stabilityiniclr2024,rectified}.

\paragraph{Decode for tails on both modalities}
Use larger decoding budgets and avoid very hard truncation for text and for images. Low temperature and very small top-$p$ or top-$k$ cut rare words and rare visual details. A bigger candidate pool is like casting a wider net so uncommon fish are not lost \cite{taleoftail}.

\paragraph{Cross modal monitoring}
Watch single side diversity for text and for images, and also watch cross modal agreement. For text, track entropy and distinct $n$-grams. And for images, track precision and recall and FID or a simple feature spread. For pairs, track CLIP style alignment and a simple gap between text and image spaces. Use simple retrieval checks like recall at $k$ to catch compressed embeddings or drift.

\paragraph{Adaptive mixing and rollback}
Lower the synthetic share $\lambda_{t}$ for the whole system or for a single modality when signals slide. Go back to the last stable snapshot when needed. You can also switch to relabeling with the frozen anchor. Prefer accumulate not replace. Keep a steady human core and add curated synthetic data step by step \cite{stabilityiniclr2024,rectified}.

\paragraph{Provenance and audits}
Log decoding settings on both sides such as temperature and top-$p$ or top-$k$ and the candidate budget. Record how each sample was produced. This is like keeping a cooking diary so you can find which step made the dish too bland and fix it. These checks mirror the single modality rules and add cross modal guards so coupled training stays in the stable zone \cite{taleoftail,stabilityiniclr2024,rectified}.

\subsection{Fidelity and Scalability Calibration}

Algorithmic countermeasures face a practical constraint. High fidelity human data provides strong anchoring signals, but it is costly and does not scale well. Methods that add entropy, protect rare samples, or adjust the synthetic share work best when there is always a minimum level of human verified data. They act like guardrails, not replacements. They help stabilize the path, but they cannot replace the solid ground beneath the model.

In practice, the level of synthetic data depends on the risk of the task.
\begin{itemize}
    \item Low risk creative generation, such as images or texts, can use a high synthetic share from $60\%$ to $90\%$ as long as entropy and diversity stay in a healthy range.
    \item Instruction following language models should keep synthetic data at a moderate level from $30\%$ to $50\%$ with tail aware weighting to avoid losing niche skills.
    \item Safety critical reasoning, such as medical, legal, or financial tasks, should cap synthetic data at $10\%$ or less since even small shifts can grow across rounds.
\end{itemize}

Together these settings form a fidelity scalability frontier. A small amount of high fidelity human data prevents severe drift, while scalable synthetic data helps expand coverage. Algorithmic tools such as entropy bands, tail weights, or simple sharpness checks can widen this frontier by slowing collapse, but they cannot replace the anchoring effect of real human data. A simple rule is to keep a non zero human core at all times, even if small, and treat synthetic augmentation as a speed boost, not the engine. Increase it only when signals remain healthy.

\subsection{Monitoring and Rollback}
Run MC risk like site reliability with clear metrics, simple thresholds, and a short playbook.
Track the following each round and as short moving averages:
\begin{enumerate}
  \item Tail coverage. Measure rare cases in text and in vision.
  \item Diversity signals. Use distinct $n$-grams and entropy for text. Use precision and recall and feature spread for images. Use CLIP style alignment and a simple modality gap for multimodal.
  \item Scaling slope. Fit log loss against log data or compute and watch for drops or bends.
\end{enumerate}

Set traffic light thresholds for each metric. Use green, amber, and red based on simple statistics, for example $z$ scores or bootstrap intervals. When a metric turns amber move to a guarded mode. When it turns red run the rollback plan from top to bottom:
\begin{enumerate}
  \item Lower the synthetic share $\lambda_t$ for the whole system or for one modality, and stop learning rate increases.
  \item Increase decoding diversity, raise temperature and the candidate budget, relax top-$p$ or top-$k$ to restore tail mass.
  \item Add or upweight a real data anchor and mix more human data this round. In multimodal loops freeze one human trained component for relabeling or calibration.
  \item Revert to the last stable checkpoint and resume with tighter monitors and a slower schedule.
\end{enumerate}

Monitoring complexity and overhead remain low because all signals are computed as local summaries rather than global scans. In practice, tail coverage and text diversity scale linearly with batch size and token count, $O(BN)$, while visual feature statistics scale with embedding dimension $O(Bd)$, consistent with their role in preserving rare modes \cite{taleoftail,modelcollapse2024}. CLIP alignment adds a single forward pass per sample and remains bounded by $O(B)$ in multimodal setups \cite{multimodalcollapse}. Scaling slope is estimated from moving averages and requires only a few regression steps over a short window, following established approaches for tracking deviations from expected scaling behavior \cite{scalinglaw}. No full dataset passes, second order curvature estimates, or re-training sweeps are needed. As a result, monitoring can be executed every $k$ steps without materially affecting throughput.

These steps keep the system in a stable region. The synthetic share remains below task limits, decoding does not cut the tail, and a steady human data drive is maintained. Together these practices help iterative self training remain locally contractive \cite{howbad,taleoftail,stabilityiniclr2024}.

\section{Conclusions and Future Directions}
\label{sec:conclusions}

In this paper, we review existing studies on the emerging phenomenon of MC, which has been observed in various scenarios, in the self-consuming loop of using generated synthetic data to recursively train a generative model. We compare MC with related phenomena, including catastrophic forgetting, subliminal learning, neural collapse, and data poisoning, clarifying both conceptual boundaries and practical implications. By examining its underlying mechanisms, such as distributional drift and recursive self-training, we highlight how generative models suffer from performance degradation, loss of diversity, and reduced trustworthiness when recursively trained on synthetic data. Furthermore, we also review mitigation strategies proposed to date, ranging from data-centric approaches to algorithmic approaches.

Overall, our analysis underscores that MC is not merely a technical issue, but a fundamental concern for the reliability and trustworthiness of GenAI. As the community increasingly relies on synthetic data, MC raises urgent questions about the long-term stability, fairness, and accountability of AI models. Based on the review, we highlight several challenges and opportunities that are worth future research efforts.

\subsection{Mitigation Strategies Beyond Real-Synthetic Balancing}
Most proposed countermeasures rely on maintaining a sufficient proportion of real data in the training set. While necessary, this strategy is not scalable, as high-quality real-world data is both limited and costly to acquire. Future research should explore complementary strategies, such as adaptive filtering of low-quality synthetic samples, self-regularizing loss functions that penalize overfitting to recursive artifacts, and hybrid pipelines that incorporate human-in-the-loop validation. Additionally, adversarial training and contrastive objectives may help preserve distributional robustness, while continual calibration against high-quality anchors (trusted datasets or golden models) could slow collapse even in low-data settings.

Another promising direction for mitigating MC lies in machine unlearning \cite{machineunlearning} and immune AI \cite{immuneai} techniques. In the context of MC, where a model may inherit undesirable artifacts or degenerate behaviors from synthetic training data, unlearning could serve as a corrective mechanism to erase the harmful knowledge while retaining useful capabilities. From the perspective of immune AI, the model training system can be made resilient to unreliable data samples such as those synthesized by generative models through immunization with digital vaccines which activate a defense mechanism to protect the trained model from collapse.

\subsection{Characterizing Synthetic Data Across Modalities}
Synthetic data is not homogeneous. Its quality and diversity vary across modalities. A deeper understanding is needed of how properties such as semantic fidelity, intra-class diversity, and latent bias accumulation affect MC. For instance, in text generation, repetitive phrasing and loss of rare linguistic constructs may accelerate collapse, while in image generation, mode collapse phenomena in GANs may amplify recursive degradation. Importantly, the speech domain remains underexplored. To the best of our knowledge, no studies have yet systematically investigated MC in speech synthesis or speech recognition tasks. Thus, exploring whether recursive training degrades prosody, phonetic diversity, or speaker generalization represents a fertile new research frontier.

\subsection{Federated and Decentralized Learning Contexts}
A particularly important and largely unexplored setting is federated learning. In decentralized training ecosystems, clients may introduce synthetic data, with or without intention, generated by local models, which could propagate collapse phenomena across the federation. This raises new challenges: How does MC manifest when updates are aggregated from distributed sources? Could recursive contamination amplify faster under non-i.i.d. data distributions? Addressing these questions requires adapting collapse analysis to the federated learning paradigm and developing aggregation rules, anomaly detection methods, and trust mechanisms that are robust to synthetic data pollution. Since federal learning is increasingly used in privacy-sensitive applications like healthcare and finance, ensuring GenAI models resilient to collapse in this context is of critical practical importance.

\subsection{Cross-Phenomenon Connections}
Although MC is distinct from catastrophic forgetting, subliminal learning, neural collapse, and data poisoning, these phenomena share conceptual and mechanistic overlaps. Future work should investigate whether collapse can be unified with these problems under a common theoretical lens, perhaps framed as distributional degradation under recursive optimization. For example, collapse may be viewed as a generative counterpart to forgetting, where previously diverse regions of the data distribution are gradually ``forgotten" due to self-reinforcement. Exploring these analogies could inspire new countermeasures adapted from other domains, such as episodic memory mechanisms from continual learning or robust filtering from adversarial defense research.

\subsection{Licensing, Provenance, and Collapse Detection}
MC complicates both licensing compliance and data provenance, as repeated use of model-generated outputs makes it harder to distinguish original data from derived content. This increases the risk of violating attribution or usage requirements and highlights the need for provenance systems that track both external sources and synthetic data. In this regard, data watermarking \cite{watermarking} may be a practical and helpful technique for identifying the source of data. In production, early detection becomes equally important. Monitoring distribution drift, output anomalies, and the proportion of synthetic data in the pipeline can provide practical signals of emerging collapse. Combining provenance tracking with continuous monitoring offers a lightweight but effective safeguard for both legal integrity and model stability.

\end{document}